\documentclass[runningheads]{llncs}
\usepackage{graphicx}
\usepackage{amsmath,amssymb} 
\usepackage{color}
\usepackage{xcolor,colortbl}

\usepackage{array}
\newcolumntype{P}[1]{>{\centering\arraybackslash}p{#1}}
\newcolumntype{M}[1]{>{\centering\arraybackslash}m{#1}}
\usepackage[width=122mm,left=12mm,paperwidth=146mm,height=193mm,top=12mm,paperheight=217mm]{geometry}

\definecolor{amber}{rgb}{1.0, 0.75, 0.0}
\definecolor{applegreen}{rgb}{0.55, 0.71, 0.0}
\definecolor{brickred}{rgb}{0.8, 0.25, 0.33}
\def\zz#1{%
\ifdim#1pt>0.9pt\cellcolor{applegreen}\else
\ifdim#1pt>0.7pt\cellcolor{yellow}\else
\ifdim#1pt>0.4pt\cellcolor{amber}\else
\cellcolor{brickred}\fi\fi\fi
#1}

\usepackage{times}
\usepackage{epsfig}
\usepackage{epstopdf}
\usepackage{cite}
\usepackage{subfig}
\usepackage{bm}
\usepackage{verbatim}
\usepackage{multirow}
\usepackage[ruled, noend]{algorithm2e}
\usepackage[noend]{algpseudocode}
\usepackage{svg}
\usepackage{wrapfig}
\usepackage[symbol]{footmisc}

\newcommand{\figLabel}{Figure~}

\newcommand{\tblLabel}{Table~}

\newcommand\mypara[1]{\vspace{4pt}\noindent\textbf{#1}}

\begin{document}
\pagestyle{headings}
\mainmatter
\def\ECCV18SubNumber{2}  

\title{Teaching UAVs to Race:\\ End-to-End Regression of Agile Controls in Simulation}

\titlerunning{Teaching UAVs to Race: End-to-End Regression of Agile Controls in Simulation}
\authorrunning{Matthias M\"uller$^{\fnsymbol{footnote}}$, Vincent Casser$^{\fnsymbol{footnote}}$, Neil Smith, Dominik L. Michels, Bernard Ghanem}

\author{Matthias M\"uller\thanks{equal contribution}, Vincent Casser$^{\fnsymbol{footnote}}$, Neil Smith, Dominik L. Michels, Bernard Ghanem\\
		{\tt\small \{matthias.mueller.2, vincent.casser, neil.smith, dominik.michels, bernard.ghanem\}@kaust.edu.sa}}

\institute{Visual Computing Center at King Abdullah University of Science and Technology}

\maketitle

\begin{abstract}
Automating the navigation of unmanned aerial vehicles (UAVs) in diverse scenarios has gained much attention in recent years. However, teaching UAVs to fly in challenging environments remains an unsolved problem, mainly due to the lack of training data. In this paper, we train a deep neural network to predict UAV controls from raw image data for the task of autonomous UAV racing in a photo-realistic simulation. Training is done through imitation learning with data augmentation to allow for the correction of navigation mistakes. Extensive experiments demonstrate that our trained network (when sufficient data augmentation is used) outperforms state-of-the-art methods and flies more consistently than many human pilots. Additionally, we show that our optimized network architecture can run in real-time on embedded hardware, allowing for efficient on-board processing critical for real-world deployment. From a broader perspective, our results underline the importance of extensive data augmentation techniques to improve robustness in end-to-end learning setups.
\end{abstract}


\section{Introduction} \label{sec:intro}
\begin{wrapfigure}{r}{0.5\textwidth}
\vspace{-24pt}
\includegraphics[width=\linewidth]{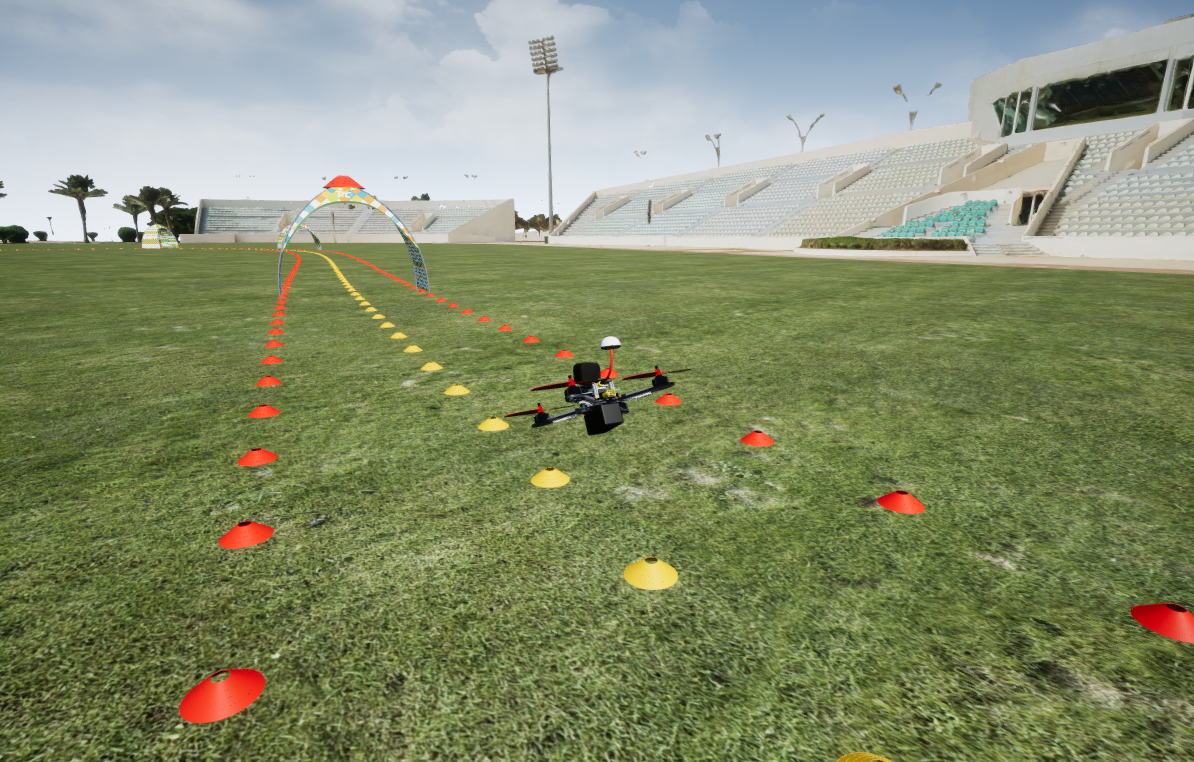}
\caption{Illustration of the trained racing UAV in-flight.}
\label{fig:Teaser}
\vspace{-20pt}
\end{wrapfigure}
Unmanned aerial vehicles (UAVs) like drones and multicopters are attracting increased interest across various communities such as robotics, graphics, and computer vision. Learning to control UAVs in complex environments is a challenging task even for humans. One of the most challenging navigation tasks with respect to UAVs is competitive drone racing. It takes extensive practice to become a good pilot, frequently involving crashes. A more affordable approach to develop professional flight skills is to train many hours in a flight simulator before going to the field. Since most of the fine motor skills of flight control are developed in the simulator, the pilot is able to quickly transition to real-world flights.

Humans are able to abstract the visual differences between simulation and the real world and are able to transfer the learned control knowledge with some finetuning to account for the small differences of the physics simulation.
While transfer for trained network policies is more difficult due to the perception component, it will be easier if the simulation is as close to reality as possible.  
Therefore, we use the physics-based UAV racing game within Sim4CV \cite{sim4cv} which features a photo-realistic and customizable racing area in the form of a stadium based on a three-dimensional (3D) scanned real-world location. This ensures minimal discrepancy when transitioning from the simulated to a real-world scenario in the future. The concept of generating synthetic clones of real-world data for deep learning purposes has been adopted in previous work \cite{gaidon2016virtual}. Also, it has become popular recently to use video game engines \cite{carla, AirSim} to generate photo-realistic simulations for training autonomous agents.

Combining the realistic physics and graphics of a game engine coupled with a real-world 3D scan should make the transfer much simpler and fine-tuning on some real world data may be sufficient if a sufficiently robust policy was trained in simulation. A key requirement for generalization is the DNN's ability to learn the appearance of gates and cones in the track within a complexly textured and dynamic environment. In the simulated environment, we have the opportunity to fully customize the race track, including using different textures (e.g. grass, snow, and dirt), gates (different shapes and appearance), and lighting. This will make the trained network more robust and will enable transfer to the real world via domain randomization \cite{Sadeghi2017}.

Our autonomous racing UAV approach goes beyond simple pattern detection and instead learns a full end-to-end system to fly the UAV through a racing course. It is similar in spirit to learning an end-to-end driving policy for a car \cite{NvidiaCar}, but comes with additional challenges. The proposed network extends the complexity of previous work to the control of a six degrees of freedom (6-DoF) flying system which is able to traverse tight spaces and make sharp turns at very high speeds (a task that cannot be performed by a ground vehicle). Our imitation learning based approach simultaneously addresses both problems of perception and control as the UAV navigates through the course. 

\mypara{Contributions.} Our specific contributions are as follows.

\textbf{(1)} We show that the challenging task of UAV racing can be learned in an end-to-end fashion in simulation, and both demonstrate and quantify the positive impact of using viewpoint augmentation for increased robustness. Experiments show that our trained network can outperform several baselines and fly more consistently than the pilots on whose data it was trained.

\textbf{(2)} To facilitate the training, parameter tuning and evaluation of deep networks on this type of simulated data, we provide a full integration between the simulator and an end-to-end deep learning pipeline (based on TensorFlow). Similar to other deep networks trained for game play, our integration will allow the community to fully explore many scenarios and tasks that go far beyond UAV racing in a rich and diverse photo-realistic gaming environment (e.g.~obstacle avoidance and trajectory planning).

\textbf{(3)} We integrate a photo-realistic UAV racing simulation environment based on a real-world counterpart which can be easily customized to build increasingly challenging racing courses and enables realistic UAV physical behavior. Logging video data from the UAV's point-of-view and pilot controls is seamless and can be used to effortlessly generate large-scale training data for AI systems targeting UAV flying in particular and autonomous vehicles in general (e.g.~self-driving cars).

\section{Related Work} \label{sec:related work}
In this section, we put our proposed methodology into context, focusing on the most related previous work.

\mypara{Learning to Navigate.} 
Navigation has traditionally been approached by either employing supervised learning (SL) methods \cite{NvidiaCar,deepDriving,ForestTrail,pomerleau1989alvinn,NIPS2005,Koutnik:2013,Dagger} or reinforcement learning (RL) methods \cite{deepReinforcementSimulator,2016-TOG-deepRL,2017-TOG-deepLoco,bikeStunts,AtariNature,mnih2016asynchronous}. Furthermore, combinations of the two have been proposed in an effort to leverage advantages of both techniques, e.g. for increasing sample efficiency for RL methods \cite{Andersson2017,Atari,guidedpolicysearch,DosovitskiyK16,parkourSimulation}. For the case of controlling physics-driven vehicles, SL can be advantageous when acquiring labeled data is not too costly or inefficient, and has been proven to have relative success in the field of autonomous driving, among other applications, in recent years \cite{NvidiaCar,deepDriving,ForestTrail}. However, the use of neural networks for SL in autonomous driving goes back to much earlier work \cite{pomerleau1989alvinn,NIPS2005}.

In the work of Bojarski et al.~\cite{NvidiaCar}, a deep neural network (DNN) is trained to map recorded camera views to 3-DoF steering commands (steering wheel angle, throttle, and brake). Seventy-two hours of human driven training data was tediously collected from a forward facing camera and augmented with two additional views to provide data for simulated drifting and corrective maneuvering.
The simulated and on-road results of this pioneering work demonstrate the ability of a DNN to learn (end-to-end) the control process of a self-driving car from raw video data.

Similar to our work but for cars, Chen et al.~\cite{deepDriving} use TORCS (The Open Racing Car Simulator) \cite{torcs} to train a DNN to drive at casual speeds through a course and properly pass or follow other vehicles in its lane. This work builds off earlier work using TORCS, which focused on keeping the car on a track \cite{Koutnik:2013}.
In contrast to our work, the vehicle controls to be predicted in the work of Chen et al.~\cite{deepDriving} are limited, since only a small discrete set of expected control outputs are available: turn-left, turn-right, throttle, and brake. Recently, TORCS has also been successfully used in several RL approaches for autonomous car driving \cite{mnih2016asynchronous, deepReinforcementSimulator,Koutník2014}; however, in these cases, RL was used to teach the agent to drive specific tracks or all available tracks rather than learning to drive never before seen tracks.

Loquercio et al.~\cite{loquercio2018dronet} trained a network on autonomous car datasets and then deployed it to control a drone. For this, they used full supervision by providing image and measured steering angle pairs from pre-collected datasets, and collecting their own dataset containing image and binary obstacle indication pairs. While they demonstrate an ability to transfer successfully to other environments, their approach does not model and exploit the full six degrees of freedom available. It also focuses on slow and safe navigation, rather than optimizing for speed as is the case for racing. Finally, with their network being fairly complex, they report an inference speed of 20fps (CPU) for remote processing, which is more than three times lower than the estimated frame rate for our proposed method when running on-board processing, and more than 27 times lower compared to our method running remotely on GPU.

In the work of Smolyanskiy et al.~\cite{ForestTrail}, a DNN is trained (in an SL fashion and from real data captured from a head-mounted camera) to navigate a UAV through forest trails and avoid obstacles. Similar to previous work, the expected control outputs of the network are discrete and very limited (simple yaw movements): turn-left, go-straight, or turn-right. Despite showing relatively promising results, the trained network  leads to a slow, non-smooth (zig-zag) trajectory at a fixed altitude above the ground. It is worthwhile to note that indoor UAV control using DNNs has also been recently explored \cite{Andersson2017,Kim2015DeepNN,Shah:2016}.

\mypara{Importance of Exploration in Supervised Learning.} In imitation learning \cite{Hussein:2017:ILS:3071073.3054912}, the `expert' training set used for SL is augmented and expanded, so as to combine the merits of both exploitation and exploration. In many sequential decision making tasks of which autonomous vehicle control is one, this augmentation becomes necessary to train an AI system (e.g.~DNN) that can recover from mistakes. In this sense, imitation learning with augmentation can be crudely seen as a supervision guided form of RL. For example, a recent imitation learning method called DAgger (Dataset Aggregation) \cite{Dagger} demonstrated a simple way of incrementally augmenting ground-truth sequential decisions to allow for further exploration, since the learner will be trained on the aggregate dataset and not only the original expert one. This method was shown to outperform state-of-the-art AI methods on a 3D car racing game (Super Tux Kart), where the control outputs are again 3-DoF. Other imitation learning approaches \cite{guidedpolicysearch} have reached a similar conclusion, namely that a trajectory optimizer can function to help guide a sub-optimal learning policy towards the optimal one. Inspired by the above work, our proposed method also exploits similar concepts for exploration. In the simulator, we are able to automatically and effortlessly generate a richly diverse set of image and control pairs that can be used to train a UAV to robustly and reliably navigate through a racing course.

\mypara{Simulation.} As mentioned earlier, generating diverse `natural' training data for sequential decision making through SL is tedious. Generating additional data for exploration purposes (i.e.~in scenarios where both input and output pairs have to be generated) is much more so. 
Therefore, a lot of attention from the community is being given to simulators (or games) for this source of data. In fact, a broad range of work has exploited them recently for these types of learning, namely in animation and motion planning \cite{birdFlightSimulator,unity3Dphysics,deepReinforcementSimulator,UE4simulator,bikeStunts,balancingSimulator,parkourSimulation}, scene understanding \cite{Battaglia05112013,syntheticRGBD}, pedestrian detection \cite{Pedestrian2010},  and identification of 2D/3D objects \cite{syntheticCarRecognition,syntheticVehicleTraining,teaching3D}. For instance, the authors of \cite{birdFlightSimulator} used Unity, a video game engine similar to Unreal Engine, to teach a bird how to fly in simulation.

Moreover, there is another line of work that uses hardware-in-the-loop (HILT) simulation. Examples include JMAVSim \cite{uavHIL2015,hilUAV} which was used to develop and evaluate controllers and RotorS \cite{Frrr2016} which was used to study visual servoing. The visual quality of most HIL simulators is very basic and far from photo-realistic with the exception of AirSim \cite{AirSim}. While there are multiple established simulators such as Realflight, Flightgear, or XPlane for simulating aerial platforms, they have several limitations. In contrast to Unreal Engine, advanced shading and post-processing settings are not available and the selection of assets and textures is limited. Recent work \cite{gaidon2016virtual,GtaV, carla, AirSim, sim4cv} highlights how modern game engines can be used to generate photo-realistic training datasets and pixel-accurate segmentation masks. The goal of this work is to build an automated UAV flying system (based on imitation learning) that can relatively easily be transitioned from a simulated world to the real one. Therefore, we choose Sim4CV ~\cite{Mueller2016, sim4cv} as our simulator, which uses the open source game engine UE4 and provides a full software in-the-loop UAV simulation. The simulator also provides a lot of flexibility in terms of assets, textures, and communication interfaces.

\section{Methodology} \label{sec:overview}
\begin{figure}[!htb]
	\includegraphics[width=\linewidth]{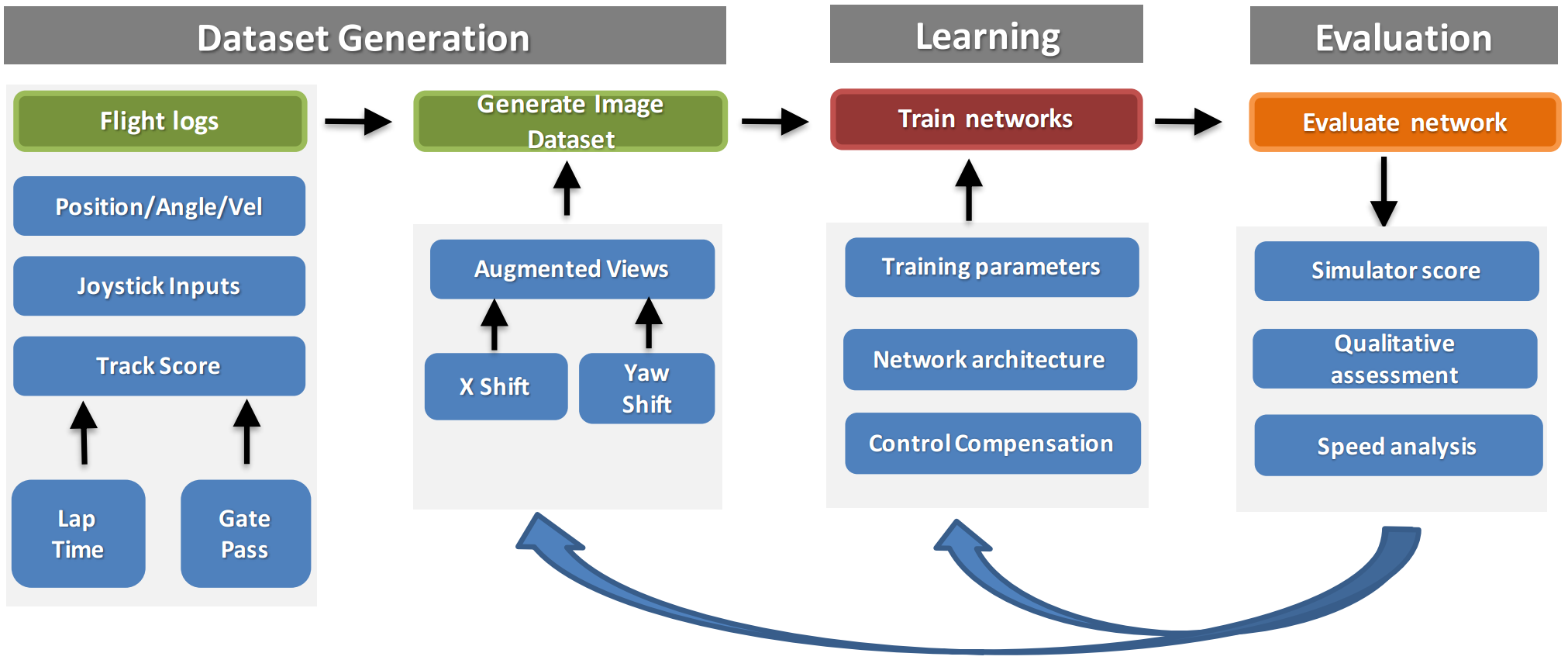}
	\caption{Description of the pipeline of our DNN Imitation Learning System. After recording flights of human pilots, we improve important model parameters like network architecture, number of augmented views and appropriate control compensation for them in an iterative process.}
	\label{fig:pipeline}
	     \vspace{-26pt}	
\end{figure}

The fundamental modules of our proposed system are summarized in \figLabel~\ref{fig:pipeline}, which represents the end-to-end dataset generation, learning, and evaluation process. In what follows, we provide details for each of these modules, namely how datasets are automatically generated within the simulator, how our proposed DNN is designed and trained, and how the learned DNN is evaluated. 

\subsection{Dataset Generation} \label{sec:dataset}
Our simulation environment allows for the automatic generation of customizable datasets that can be used for various learning tasks related to UAVs. In the following, we elaborate on our setup for building a large-scale dataset specific to UAV racing.

\begin{wrapfigure}{r}{0.5\textwidth}
\vspace{-16pt}
   \includegraphics[width=\linewidth]{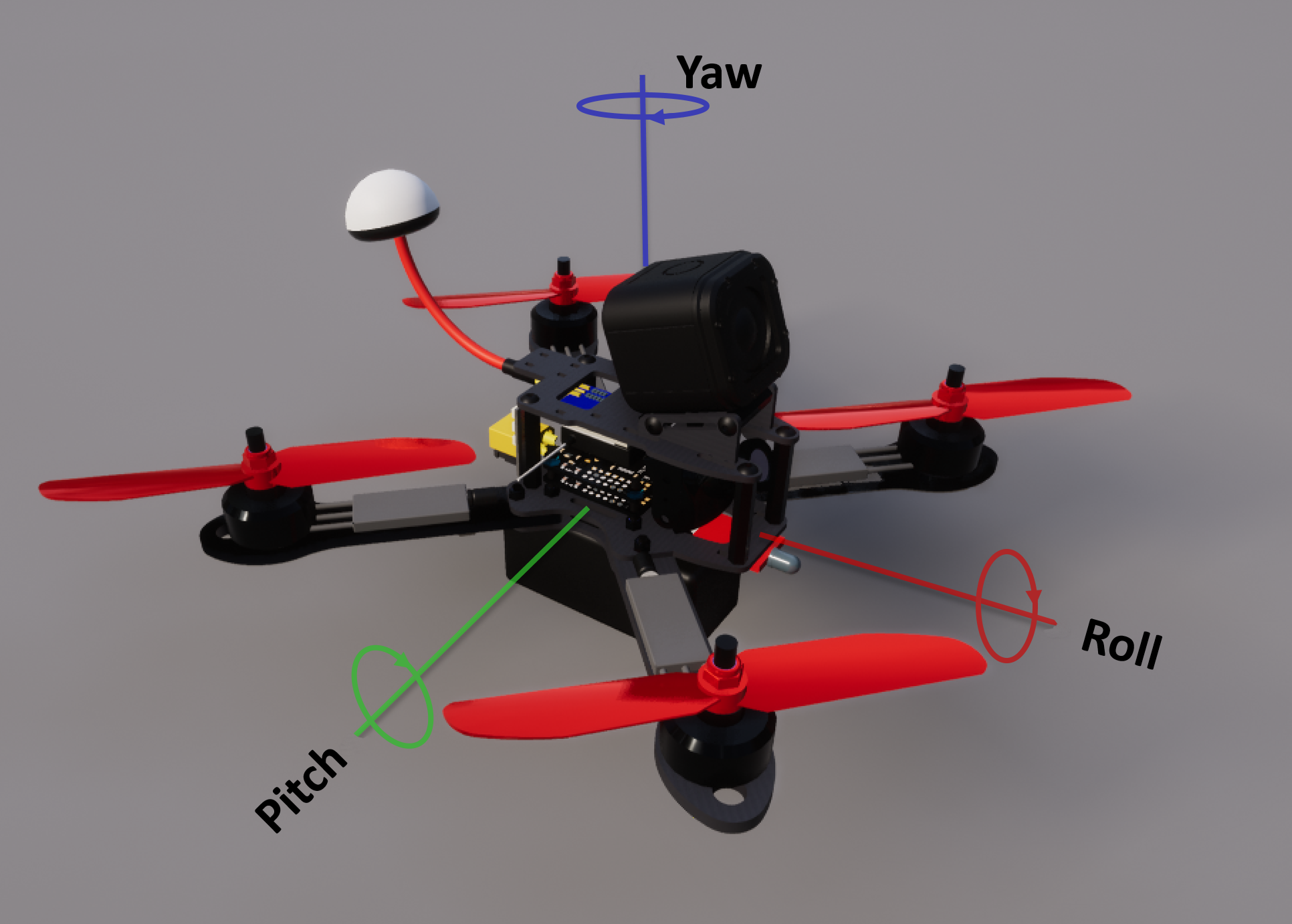}
     \caption{The 3D model of the racing UAV modeled in the simulator, based on a well known 250 class design known within the racing community as the \textit{Hornet}.}\label{fig:uav_attitude}
     \vspace{-20pt}	
\end{wrapfigure}

\mypara{UAV Flight Simulation.} The core of the system is the application of our UE4 based simulator. It is built on top of the open source UE4 project for computer vision called Sim4CV \cite{sim4cv}. Several changes were made to adapt the simulator for training our proposed racing DNN. First, we replaced the UAV with the 3D model and specifications of a racing quadcopter (see \figLabel \ref{fig:uav_attitude}). We retuned the PID controller of the UAV to be more responsive and to function in a racing mode, where altitude control and stablization are still enabled but with much higher rates and steeper pitch and roll angles. In fact, this is now a popular racing mode available on consumer UAVs, such as the DJI Mavic. The simulator frame rate is locked at 60\,fps and at every frame a log is recorded with UAV position, orientation, velocity, and stick inputs from the pilot. To accommodate for realistic input, we integrated the same UAV transmitter that would be used in real-world racing scenarios. 

Following paradigms set by UAV racing norms, each racing course/track in our simulator comprises a sequence of gates connected by uniformly spaced cones.  The track has a timing system that records time between each gate, lap, and completion time of the race. The gates have their own logic to detect whether the UAV has passed through the gate in the correct direction. This allows us to trigger both the start and ending of the race, as well as, determine the number of gates traversed by the UAV. These metrics (time and percentage of gates passed) constitute the overall per-track performance of a pilot, be it a human or a DNN.

\begin{figure*}[!htb]
  \centering
    \vspace{-12pt}
  \includegraphics[width=0.99\textwidth]{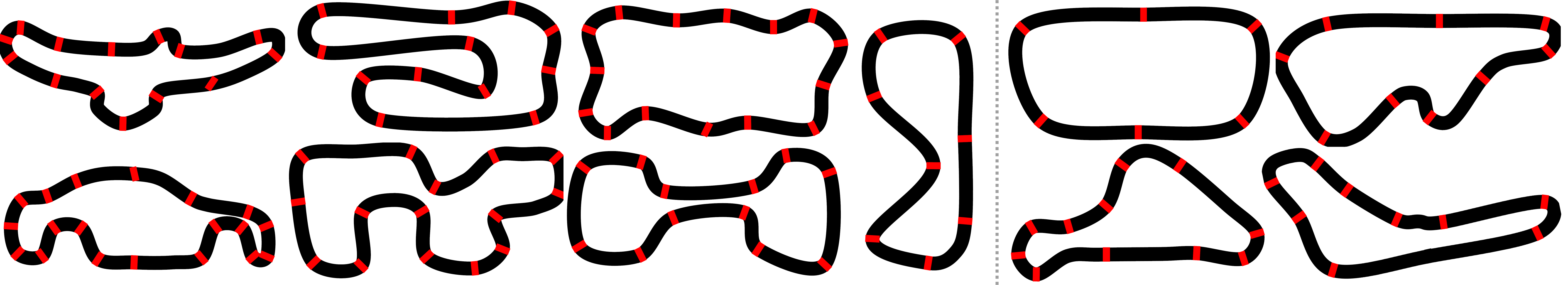}
	  \caption{The seven training tracks (left) and the four evaluation tracks (right). Gates are marked in red.}
		\label{fig:testtracks}
    \vspace{-16pt}
\end{figure*}

\mypara{Automatic Track Generation.} We developed a graphical track editor in which a user can draw a 2D sketch of the overhead view of the track. Subsequently, the 3D track is automatically generated and integrated into the timing system. With this editor, we created eleven tracks: seven for training, and four for testing and evaluation. Each track is defined by gate positions and track lanes delineated by uniformly spaced racing cones distributed along the splines connecting adjacent gates. We design the tracks such that they are similar to what racing professionals are accustomed to and such that they offer enough diversity to enable network generalization to unseen tracks. To avoid user bias in designing the race tracks, we use images collected from the internet and trace their contours in the editor to create uniquely stylized tracks. Please refer to \figLabel~\ref{fig:testtracks} for an overhead view of all these tracks.

\mypara{Acquiring Ground-truth Pilot Data.} The simulation environment allows us to log the images rendered from the UAV camera point-of-view and the UAV flight controls from the transmitter. We record human pilot input from a Taranis flight transmitter integrated into the simulator through a joystick. This input is solicited from three pilots with different skill levels: novice (lacking any flight experience), intermediate (a moderately experienced pilot), and expert (a professional, competitive racing pilot). The pilots are given the opportunity to fly through the seven training tracks as many times as needed until they successfully complete the tracks at their best time while passing through all gates. For the evaluation tracks, the pilots are allowed to fly the course only as many times as needed to complete the entire course without crashing. We automatically score pilot performance based on lap time and percentage of gates traversed. 

\mypara{Data Augmentation.}
As mentioned earlier, robust imitation learning requires the augmentation of these ground-truth logs with synthetic ones generated at a user-defined set of UAV offset positions and orientations accompanied by the corresponding controls needed to correct for these offsets. Assigning corrective controls to the augmented data is quite complex in general, since they depend on many factors, including current UAV velocity, relative position on the track, its weight and current attitude. While it is possible to get this data in the simulation, it is very difficult to obtain it in the real world in real-time. Therefore, we employ a fairly simple but effective model to determine these augmented controls that also scales to real-world settings. We add or subtract a corrective value to the pilot roll and yaw stick inputs for each position or orientation offset that is applied. For rotational offsets, we do not only apply a yaw correction but also couple it to roll. This allows to compensate for the UAV's inertia which produces a motion component in the previous direction of travel.

\begin{wraptable}{r}{0.55\textwidth}
\centering
    \vspace{-20pt}
	\small\addtolength{\tabcolsep}{4pt}
	\begin{tabular}{l|c|c|r}
    \hline
    track & duration (sec) & original & total \\ \hline
    track01 & 69.8& 4.2K & 29.3K \\
	track02 & 100.4 & 6.0K & 42.2K \\
	track03 & 83.1 & 5.0K & 35.0K \\
	track04 & 97.7 & 5.9K & 41.0K \\
	track05 & 99.8 & 6.0K & 42.0K \\
	track06 & 115.4 & 6.9K & 48.5K \\
	track07 & 98.3 & 5.9K & 41.2K \\
	total & 664.5 & 39.9K & 279.1K \\ \hline		
  \end{tabular}
  \caption{Overview of the image-control dataset generated from two laps of flying (by the intermediate pilot) through each of the training tracks. The `duration' column shows the total time taken by the pilot to successfully fly two laps through the track (i.e. passing through all the gates). We also record the number of images rendered from the pilot's trajectory in the simulator, along with the total number of images used for training when data augmentation is applied. For this augmentation, we use the following default settings: roll offset ($\pm 50$cm), yaw offset ($\pm 15^{\circ}$ and $\pm 30^{\circ}$).}
  \label{tbl:dataset}
  \vspace{-20pt}	
\end{wraptable}

\mypara{Training Data Set.}
We summarize the details of the data generated from all training tracks in Table \ref{tbl:dataset}. It is clear that the augmentation increases the size of the original dataset by approximately seven times.
Each pilot flight leads to a large number of image-control pairs (both original and augmented) that will be used to train the UAV to robustly recover from possible drift along each training track, as well as, generalize to unseen evaluation tracks. Details of how our proposed DNN architecture is designed and trained are provided in Section 3.2 of the paper. In general, more augmented data should improve UAV flight performance assuming that the control mapping and original flight data are noise-free. However, in many scenarios, this is not the case, so we find that there is a limit after which augmentation does not help (or even slightly degrades) explorative learning. Empirical results validating this observation are detailed in Section 4 of the paper. We also show the effects of training with different flying styles there. For this dataset, we choose to use the intermediate pilot who tends to follow the track most precisely, striking a good trade-off between style of flight and speed. 

Since the logs can be replayed at a later time in the simulator, we can augment the dataset further by changing environmental conditions, including lighting, cone spacing or appearance, and other environmental dynamics (e.g.~clouds), but we do not explore these capabilities in this work.

\subsection{Learning} \label{sec:learning}
As it is the case for DNN-based solutions to other tasks, a careful construction of the training set is a key requirement to robust and effective DNN training. We dedicate seven racing tracks with their corresponding image-control pairs logged from human pilot runs and appropriate augmentation for training. Please refer to Section \ref{sec:dataset} for details about data collection and augmentation. In the following, we provide a detailed description of the learning strategy used to train our DNN, its network architecture and design. We also explore some of the inner workings of one of the trained DNNs to shed light on how this network is solving the problem of automated UAV racing.

\begin{figure}[!htb]
  \centering
  \includegraphics[width=0.95\linewidth]{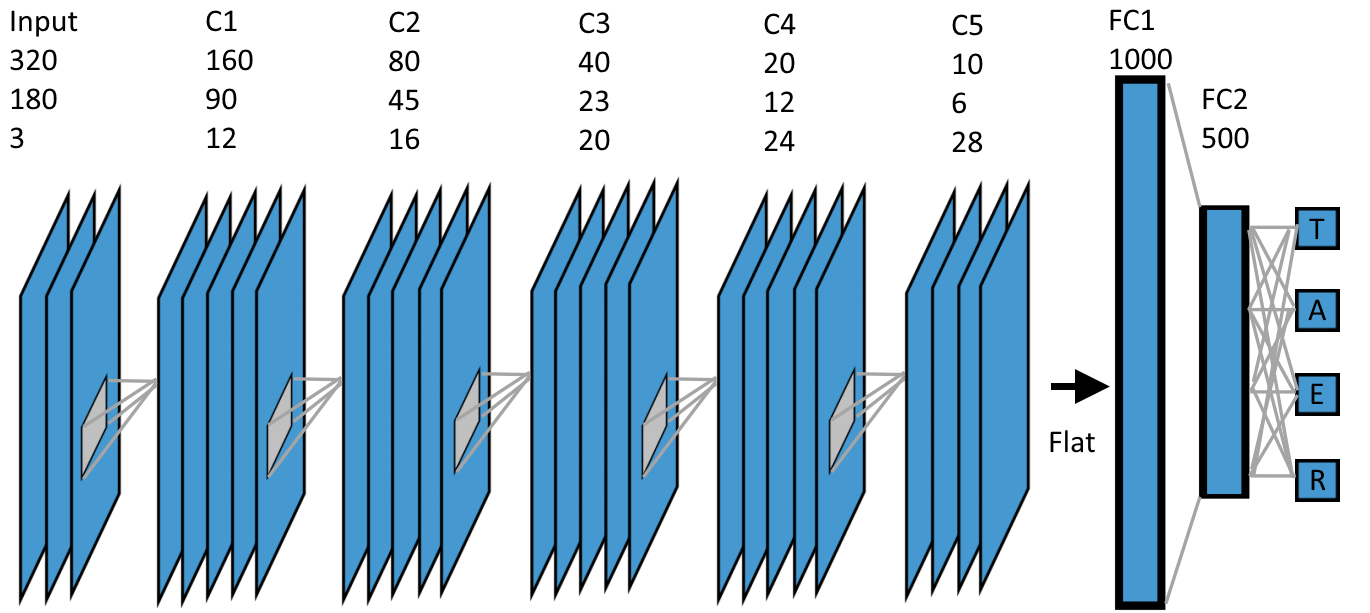}
	  \caption{Our network architecture is taking an image of shape 320x180 and regresses to the control outputs throttle (T), elevator (E), aileron (A) and roll (R).}
		\label{fig:networkarchitecture}
\end{figure}

\mypara{Network Architecture.}
To train a DNN to predict stick inputs controlling the UAV from images, we choose a regression network architecture similar to the one used by Bojarski et al.~\cite{NvidiaCar}; however, we make changes to accommodate the complexity of the task at hand and to improve robustness in training. Our DNN architecture is shown in \figLabel \ref{fig:networkarchitecture}. The network consists of eight layers, five convolutional and three fully-connected. Since we implicitly want to localize the track and gates, we use striding in the convolutional layers instead of (max) pooling.

We arrived at this compact network architecture by running extensive validation experiments. Our final architecture strikes a reasonable tradeoff between computational complexity and predictive performance.
This careful design makes the proposed DNN architecture feasible for real-time applications on embedded hardware (e.g. Nvidia~TX1, or the recent Nvidia~TX2) unlike previous architectures \cite{NvidiaCar}, if they use the same input size. In Table \ref{fig:tablegpus}, we show both evaluation time on and technical details of the NVIDIA Titan X, and how it compares to a NVIDIA TX-1. Based on \cite{tx1benchmark}, we expect our network to still run at real-time speed with over 60 frames per second on this embedded hardware.

\begin{table}
    \centering
    \begin{tabular}{l|c|r}
    \hline
                      & NVIDIA Titan X & NVIDIA TX-1 \\ \hline
    CUDA cores        & 3,840          & 256         \\
    Boost Clock MHz   & 1,582          & 998         \\
    VRAM              & 12 GB          & 4 GB        \\
    Memory Bandwidth  & 547.7 Gbps     & 25.6 Gbps   \\
    Evaluation (ours) & 556 fps (ref)  & 64.6 fps    \\ \hline
    \end{tabular}
	\caption{Comparison of the NVIDIA Titan X and the NVIDIA TX-1. The performance of the TX-1 is approximated according to \cite{tx1benchmark}.}
	\label{fig:tablegpus}
	\vspace{-20pt}
\end{table}

\mypara{Implementation Details.}
The DNN is given a single RGB-image with a 320$\times$180 pixel resolution as input and is trained to regress to the four stick inputs to control the UAV using a standard L$^2$-loss and dropout ratio of 0.5.

We find that the relatively high input resolution (i.e.~higher network capacity), as compared to related methods \cite{NvidiaCar,ForestTrail}, is useful to learn this more complicated maneuvering task and to enhance the network's ability to look further ahead. This affords the network with more robustness needed for long-term trajectory stability. On the other hand, we found no noticeably gain when training on even higher resolutions during initial experiments. At our proposed resolution, our network still shows real-time capabilities even when being deployed on-board (Table \ref{fig:tablegpus}), marking a convincing solution to the resolution-speed trade-off. For training, we exploit a standard stochastic gradient descent (SGD) optimization strategy (namely Adam) in Tensorflow. As such, one instance of our DNN can be trained to convergence on our dataset in less than two hours on a single GPU. 

In contrast to other work where the frame rate is sampled down to 10\,fps or lower \cite{deepDriving,NvidiaCar,ForestTrail}, our racing environment is highly dynamic (with tight turns, high speed, and low inertia of the UAV), so we use a frame rate of 60\,fps. This allows the UAV to be very responsive and move at high speeds, while maintaining a level of smoothness in controls. An alternative approach for temporally smooth controls is to include historic data in the training process (e.g.~add the previous controls as input to the DNN). This can make the network more complex, harder to train, and less responsive in the highly dynamic racing environment, where many time critical decisions have to be made within a couple of frames (about 30\,ms). Therefore, we find the high learning frame rate of 60\,fps a good trade-off between smooth controls and responsiveness.

\mypara{Network Visualization.}\\
After training our DNN to convergence, we visualize how parts of the network behave. 
\begin{wrapfigure}{r}{0.5\textwidth}
	\includegraphics[width=\linewidth]{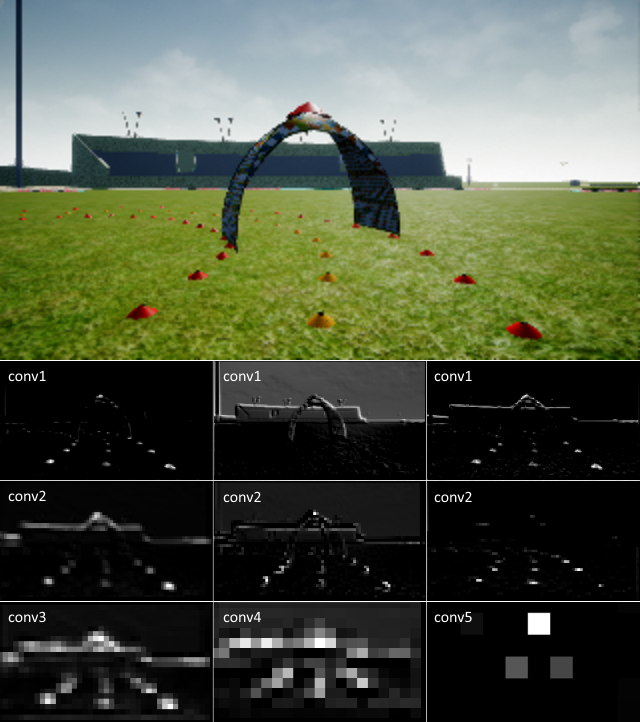}
    \caption{Visualization of feature maps at different convolutional layers in our trained network. Note the high activations in semantically meaningful image regions for the task of UAV racing, namely the gates and cones.}
    \label{fig:featuremaps}
    \vspace{-20pt}
\end{wrapfigure}
\figLabel~\ref{fig:featuremaps} shows some feature maps in different layers for the same input image. Note how the filters have learned to extract all necessary information in the scene (i.e.~gates and cones). Also, higher-level filters are not responding to other parts of the environment. Although the feature map resolution becomes very low in the higher DNN layers, the feature map in the fifth convolutional layer is interesting as it marks the top, left, and right of parts of a gate with just a single activation each. This clearly demonstrates that our DNN is learning semantically intuitive features for the task of UAV racing.

\mypara{Reinforcement vs. Imitation Learning.}
Our simulation environment can lend itself useful in training networks using reinforcement learning. This type of learning does not specifically require supervised pilot information, as it searches for an optimal policy that leads to the highest eventual reward (e.g.~highest percentage of gates traversed or lowest lap time). Recent methods have made use of reinforcement to learn simpler tasks without supervision \cite{DosovitskiyK16}; however, they require very long training times (up to several weeks) and a much faster simulator (1,000fps is possible in simple non photo-realistic games). For UAV racing, the required task is more involved and since the intent is to transfer the learned network into the real-world, a (slower) photo-realistic simulator is mandatory. Because of these two constraints, we decided to train our DNN using imitation learning instead of reinforcement learning.

\section{Experiments}\label{sec:evaluation}
We create four testing tracks based on well-known race tracks found in TORCS and Gran Turismo. We refer to \figLabel~\ref{fig:testtracks} for an overhead view of these tracks. Since the tracks must fit within the football stadium environment, they are scaled down leading to much sharper turns and shorter straight-aways with the UAV reaching top speeds of over 100\,km/h. Therefore, the evaluation tracks are significantly more difficult than they may have been originally intended in their original racing environments.
We rank the four tracks in terms of difficulty ranging from easy (track 1), medium (track 2), hard (track 3), to very hard (track 4). For all the following evaluations, both the trained networks and human pilots are tasked to fly two laps in the testing tracks and are scored based on the total gates they fly through and overall lap time. Obviously, the testing/evaluation tracks are never seen in training, neither by the human pilot nor the DNN.

\mypara{Experimental Setup.}
In order to evaluate the performance of a trained DNN in real-time at 60\,fps, we establish a TCP socket connection between the UE4 simulator and the Python wrapper (TensorFlow) executing the DNN. In doing so, the simulator continuously sends rendered UAV camera images across TCP to the DNN, which in turn processes each image individually to predict the next UAV stick inputs (flight controls) that are fed back to the UAV in the simulator using the same connection. Another advantage of this TCP connection is that the DNN prediction can be run on a separate system than the one running the simulator.
We expect that this versatile and multi-purpose interface between the simulator and DNN framework will enable opportunities for the research community to further develop DNN solutions to not only the task of automated UAV navigation (using imitation learning) but to the more general task of vehicle maneuvering and obstacle avoidance (possibly using other forms of learning including RL).

\begin{table}
\small
\centering
	\small\addtolength{\tabcolsep}{2pt}
	\begin{tabular}{|M{16mm}|M{12mm}|c|c|c|c|c|c|c|c|c|c|}
	     \cline{2-12}
    \multicolumn{1}{c|}{ } & \textbf{yaw [$^{\circ}$]} & \multicolumn{2}{|c|}{[None]} & \multicolumn{2}{|c|}{[-20:20:20]} & \multicolumn{2}{|c|}{[-30:15:30]} & \multicolumn{2}{|c|}{[-30:10:30]} & \multicolumn{2}{|c|}{[-30:5:30]} \\
   
    \hline
	 \textbf{roll [cm]} & \textbf{cameras} & \multicolumn{2}{|c|}{0} & \multicolumn{2}{|c|}{2} & \multicolumn{2}{|c|}{4} & \multicolumn{2}{|c|}{6} & \multicolumn{2}{|c|}{12} \\

    \hline
    [-75:25:75] & 6 & \zz{0.17} & \zz{0.45} & \zz{1.00} & \zz{1.00} & \zz{1.00} & \zz{1.00} & \zz{0.83} & \zz{0.85} & \zz{0.92} & \zz{1.00} \\
    \cline{3-12}
      &   & \zz{0.82} & \zz{0.50} & \zz{0.95} & \zz{1.00} & \zz{1.00} & \zz{1.00} & \zz{0.95} & \zz{1.00} & \zz{1.00} & \zz{1.00} \\
    \hline
    [-75:50:75] & 4 & \zz{0.42} & \zz{0.60} & \zz{1.00} & \zz{1.00} & \zz{1.00} & \zz{1.00} & \zz{0.75} & \zz{1.00} & \zz{1.00} & \zz{0.85} \\
    \cline{3-12}
      &   & \zz{0.82} & \zz{0.61} & \zz{0.41} & \zz{0.78} & \zz{1.00} & \zz{0.94} & \zz{0.91} & \zz{0.94} & \zz{1.00} & \zz{1.00} \\
    \hline
    [-50:50:50] & 2 & \zz{0.17} & \zz{0.35} & \zz{0.92} & \zz{1.00} & \zz{1.00} & \zz{1.00} & \zz{1.00} & \zz{1.00} & \zz{1.00} & \zz{1.00} \\
    \cline{3-12}
      &   & \zz{0.23} & \zz{0.28} & \zz{1.00} & \zz{1.00} & \zz{1.00} & \zz{1.00} & \zz{1.00} & \zz{1.00} & \zz{0.82} & \zz{1.00} \\
    \hline
    [None] & 0 & \zz{0.00} & \zz{0.00} & \zz{0.92} & \zz{1.00} & \zz{0.67} & \zz{1.00} & \zz{1.00} & \zz{1.00} & \zz{0.50} & \zz{1.00} \\
    \cline{3-12}
      &   & \zz{0.00} & \zz{0.00} & \zz{0.55} & \zz{0.78} & \zz{0.73} & \zz{1.00} & \zz{0.77} & \zz{0.89} & \zz{0.91} & \zz{0.89} \\
    \hline
  \end{tabular}
  \vspace{2mm}
  \caption{Effect of data augmentation in training to overall UAV racing performance. By augmenting the original flight logs with data captured at more  offsets (roll and yaw) from the original trajectory along with their corresponding corrective controls, our UAV DNN can learn to traverse almost all the gates of the testing tracks, since it has learned to correct for exploratory maneuvers. We show the settings abbreviated as [min:increment:max] intervals. After a sufficient amount of augmentation, no additional benefit is realized in improved racing performance.}
  \vspace{-26pt}
    \label{fig:augmentation}

\end{table}

\mypara{Effects of Exploration.}
We find exploration to be the predominant factor influencing network performance. As mentioned earlier, we augment the pilot flight data with offsets and corresponding corrective controls. We conduct grid search to find a suitable degree of augmentation and to analyze the effect it has on overall UAV racing performance. To do this, we define two sets of offset parameters: one that acts as a horizontal offset (roll-offset) and one that acts as a rotational offset (yaw-offset). \tblLabel\ref{fig:augmentation} shows how the racing accuracy (percentage of gates traversed) varies with different sets of these augmentation offsets across the four testing tracks. It is clear that increasing the number of rendered images with yaw-offset has the greatest impact on performance. While it is possible for the DNN to complete tracks without being trained on roll-offsets, this is not the case for yaw-offsets. However, the significant gain in adding rotated camera views saturates quickly, and at a certain point the network does not benefit from more extensive augmentation. Therefore, we found four yaw-offsets to be sufficient. Including camera views with horizontal shifts is also beneficial, since the network is better equipped to recover once it is about to leave the track on straights. We found two roll-offsets to be sufficient to ensure this. In the rest of our experiments, we use the following augmentation setup in training: horizontal roll-offset set $\{-50^{\circ},50^{\circ}\}$ and rotational yaw-offset set $\{-30^{\circ},-15^{\circ},15^{\circ},30^{\circ}\}$. 

\mypara{Comparison to State-of-the-Art.}
We compare our racing DNN to the two most related and recent network architectures, the first denoted as Nvidia (for self-driving cars \cite{NvidiaCar}) and the second as MAV (for forest path navigating UAVs \cite{ForestTrail}). While the domains of these works are similar, it should be noted that flying a high-speed racing UAV is a particularly challenging task, especially since the effect of inertia is much more significant and there are more degrees of freedom. For fair comparison, we scale our dataset to the same input dimensionality and re-train each of the three networks. We then evaluate each of the trained models on the task of UAV racing in the testing tracks. It is noteworthy that both the Nvidia and MAV networks (in their original implementation) use data augmentation as well, so when training, we assume the augmentation choice to be appropriate for the given method and maintain the same strategy. While the exact angular offsets of the two views used in the Nvidia network are not reported, we assume them to be close to $30^{\circ}$. We thus employ a rotational offset set of $\{-30^{\circ},30^{\circ}\}$ to augment its data. As for the MAV network, we use the same augmentation parameters proposed in the paper, i.e.~a rotational offset of $\{-30^{\circ},30^{\circ}\}$. We modified the MAV network to allow for a regression output instead of its original classification (left, center and right controls). This is necessary since our task requires fine-grained controls, and predicting discrete controls leads to very inadequate UAV racing performance.

\begin{wraptable}{r}{0.65\textwidth}
\vspace{-20pt}
\small\addtolength{\tabcolsep}{0pt}
	\begin{tabular}{l|c|c|c|c}
		\hline
		\textbf{Pilot / Network} & \textbf{Track 1} & \textbf{Track 2} & \textbf{Track 3} & \textbf{Track 4} \\ \hline
		Human-Novice & 1.00 & 1.00 & 0.95 & 0.94\\
		Human-Intermediate & 1.00 & 1.00 & 1.00 & 1.00 \\
		Human-Expert & 1.00 & 1.00 & 1.00 & 1.00 \\ \hline
		Ours-Intermediate & \textbf{1.00} & \textbf{1.00} & \textbf{1.00} & \textbf{1.00} \\
		Ours-Expert & 1.00 & 0.95 & 0.91 & 0.78 \\ \hline
		Nvidia-Intermediate & 0.17 & 1.00 & 0.82 & 0.83 \\
		Nvidia-Intermediate++ & 1.00 & 1.00 & 0.82 & 1.00 \\ \hline
		MAV-Intermediate & 0.50 & 0.75 & 0.73 & 0.83 \\
		MAV-Intermediate++ & 0.42 & 1.00 & 0.91& 0.78 \\
		\hline
	\end{tabular}
	\caption{Accuracy score of different pilots and networks on the four test tracks, averaged over multiple runs. The accuracy score represents the percentage of completed racing gates. The networks ending with ++ are variants of the original network with our augmentation strategy.}
	\label{tbl:tablecomparison}
    \vspace{-20pt}
\end{wraptable}

It should be noted that in the original implementation of the Nvidia network \cite{NvidiaCar} (based on real-world driving data), it was realized that additional augmentation was needed for reasonable automatic driving performance \emph{after} the real-world data was acquired. To avoid recapturing the data again, synthetic viewpoints (generated by interpolation) were used to augment the training dataset, which introduced undesirable distortions. By using our simulator, we are able to extract any number of camera views without distortions. Therefore, we wanted to also gauge the effect of \emph{additional} augmentation to both the Nvidia and MAV networks, when they are trained using our default augmentation setting: horizontal roll-offset of $\{-50^{\circ},50^{\circ}\}$ and rotational yaw-offset of $\{-30^{\circ},-15^{\circ},15^{\circ},30^{\circ}\}$. We denote these trained networks as Nvidia++ and MAV++.

Table~\ref{tbl:tablecomparison} summarizes the results of these different network variants on the testing tracks. Results indicate that the performance of the original Nvidia and MAV networks suffers from insufficient data augmentation. They clearly do not make use of enough exploration. These networks improve in performance when our proposed data augmentation scheme is used. Regardless, our proposed DNN outperforms the Nvidia and MAV networks, where this improvement is less significant when more data augmentation or more exploratory behavior is learned. Unlike the other networks, our DNN performs consistently well on all the unseen tracks, owing to its sufficient network capacity needed to learn this complex task.

\begin{figure*}[!htb]
	\includegraphics[width=\textwidth]{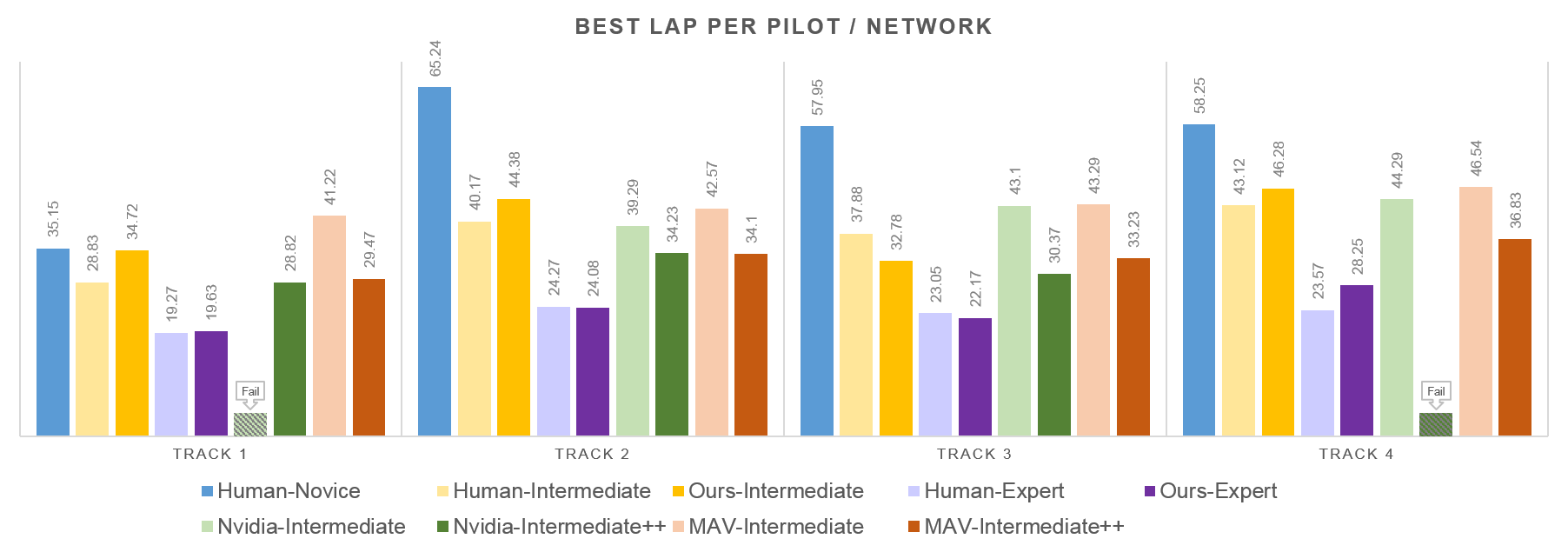}
	\caption{Best lap times of human pilots and networks trained on different flight styles. If there is no lap time displayed, the pilot was not able to complete the course because the UAV crashed. See text for a more detailed description.}
	\label{fig:pilot_diversity}
	\vspace{-12pt}
\end{figure*}

\mypara{Pilot Diversity \& Human vs. DNN.}\label{sec:pilot_diversity}

In this section, we investigate how the flying style of a pilot affects the network that is being learned. To this end, we compare the performance of the different networks on the testing set, when each of them is trained with flight data captured from pilots of varying flight expertise (intermediate and expert).

Table~\ref{tbl:tablecomparison} summarizes the lap time and accuracy of these networks. 
Clearly, the pilot flight style can significantly affect the performance of the learned network. 
\figLabel~\ref{fig:pilot_diversity} shows that there is a high correlation regarding both performance and flying style of the pilot used in training and the corresponding learned network.

The trained networks clearly resemble the flying style and also the proficiency of their human trainers. Thus, our network that was trained on flights of the intermediate pilot achieves high accuracy but is quite slow, just as the expert network sometimes misses gates but achieves very good lap and overall times. 

Interestingly, although the networks perform similar to their pilot, they fly more consistently, and therefore tend to outperform the human pilot with regards to overall time on multiple laps. This is especially true for our intermediate network. Both the intermediate and the expert network clearly outperform the novice human pilot who takes several hours of practice and several attempts to reach similar performance to the network. Even our expert pilots were not always able to complete the test tracks on the first attempt.

\begin{wrapfigure}{r}{0.5\textwidth}
	\includegraphics[width=\linewidth]{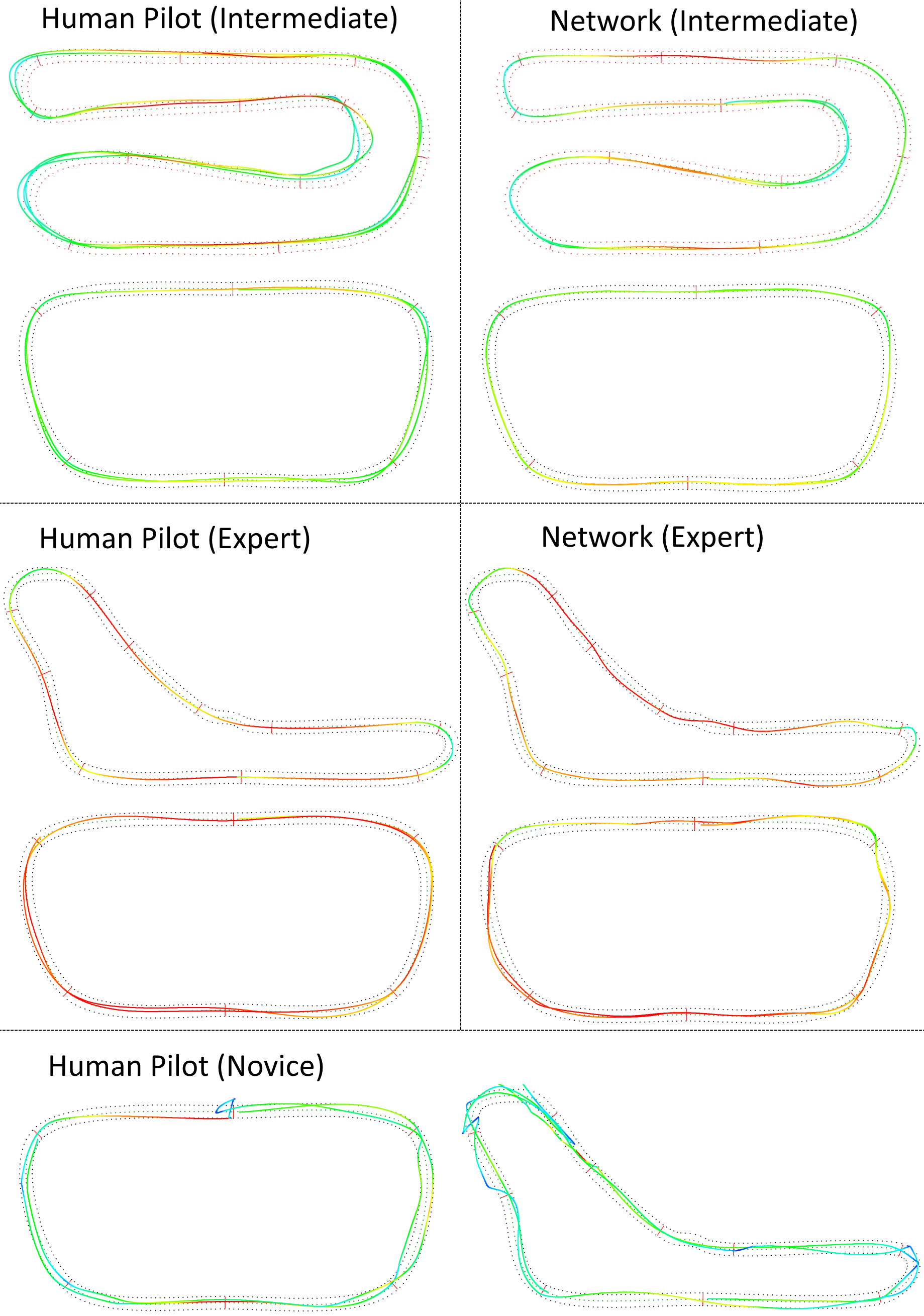}
	\caption{Visualization of human and automated UAV flights super-imposed onto a 2D overhead view of different tracks. The color illustrates the instantaneous speed of the UAV from blue (slow) to red (fast).} 
	\label{fig:course_heatmap}
	\vspace{-20pt}
\end{wrapfigure}

While the percentage of passed gates and best lap time give a good indication about the performance, they do not convey any information about the style of the pilot. To this end, we visualize the performance of human pilots and the trained networks by plotting their trajectories onto the track (from a 2D overhead viewpoint). We encode their speeds as a heatmap, where blue corresponds to the minimum speed and red to the maximum speed. \figLabel~\ref{fig:course_heatmap} shows a collection of heatmaps revealing several interesting insights. 

Firstly, despite showing variation, the networks clearly imitate the style of the pilot they were trained on. This is especially true for the intermediate proficiency level, while the expert network sometimes overshoots, which causes it to loose speed and therefore to not match the speed pattern as well as the intermediate one. We also note that the performance gap between network and human increases as the expertise of the pilot increases. Note that the flight path of the expert network is less smooth and centered than its human correspondent and the intermediate network, respectively. This is partly due to the fact that the networks were only trained on two laps of flying across seven training tracks. An expert pilot has a lot more training than that and is therefore able to generalize much better to unseen environments. 

However, the experience advantage of the intermediate pilot over the network is much less and therefore the performance gap is smaller. We also show the performance of our novice pilot on these tracks. While the intermediate pilots accelerate on straights, the novice is not able to control speed that well, creating a very narrow velocity range. Albeit flying quite slowly, the pilot also gets off track several times. This underlines the complexity of UAV racing, especially for inexperienced pilots.

\section{Conclusions and Future Work}\label{sec:conclusion}
In this paper, we proposed a robust imitation learning based framework to teach an unmanned aerial vehicle (UAV) to fly through challenging racing tracks at very high speeds. To do this, we trained a deep neural network (DNN) to predict the necessary UAV controls from raw image data, grounded in a photo-realistic simulator that also allows for realistic UAV physics. Training is made possible by logging data (rendered images from the UAV and stick controls) from human pilot flights, while they maneuver the UAV through racing tracks. This data is augmented with sufficient camera view offsets to teach the network how to recover from flight mistakes, which proves to be crucial during long-term flight. Extensive experiments demonstrate that our trained network (when such sufficient data augmentation is used) outperforms state-of-the-art methods and flies more consistently than many human pilots.

In the future, we aim to transfer the network we trained in our simulator to the real-world to compete against human pilots in real-world racing scenarios. Although we accurately modeled the simulated racing environment, the differences in appearance between the simulated and real-world will need to be reconciled. Therefore, we will investigate deep transfer learning techniques to enable a smooth transition between simulator and the real-world. If such transfer would be successful, our simulator would be able to act as an unlimited, highly customizable and free source of ground truth data.

Despite our findings that temporally aware architectures were not a good choice for the low-latency UAV racing task, we expect this to be useful when approaching general UAV navigation and complex obstacle avoidance. We plan to more broadly evaluate our method and the choice of augmentation strategy on tasks with differing challenges. More generally, since our developed simulator and its seamless interface to deep learning platforms is generic in nature, we expect that this combination will open up unique opportunities for the community to develop better automated UAV flying methods, to expand its reach to other fields of autonomous navigation such as self-driving cars, and to benefit other interesting perception-based tasks such as obstacle avoidance.

\mypara{Acknowledgments.} This work was supported by the King Abdullah University of Science and Technology (KAUST) Office of Sponsored Research through the Visual Computing Center (VCC) funding.

\clearpage

\bibliographystyle{splncs04}
\bibliography{references}

\begin{thebibliography}{10}
\providecommand{\url}[1]{\texttt{#1}}
\providecommand{\urlprefix}{URL }
\providecommand{\doi}[1]{https://doi.org/#1}

\bibitem{Andersson2017}
Andersson, O., Wzorek, M., Doherty, P.: Deep learning quadcopter control via
  risk-aware active learning. In: Thirty-First AAAI Conference on Artificial
  Intelligence (AAAI), 2017, San Francisco, February 4{-}9. : (2017), accepted.

\bibitem{Battaglia05112013}
Battaglia, P.W., Hamrick, J.B., Tenenbaum, J.B.: Simulation as an engine of
  physical scene understanding. Proceedings of the National Academy of Sciences
   \textbf{110}(45),  18327--18332 (2013). \doi{10.1073/pnas.1306572110},
  \url{http://www.pnas.org/content/110/45/18327.abstract}

\bibitem{NvidiaCar}
Bojarski, M., Testa, D.D., Dworakowski, D., Firner, B., Flepp, B., Goyal, P.,
  Jackel, L.D., Monfort, M., Muller, U., Zhang, J., Zhang, X., Zhao, J., Zieba,
  K.: End to end learning for self-driving cars. CoRR  \textbf{abs/1604.07316}
  (2016), \url{http://arxiv.org/abs/1604.07316}

\bibitem{deepDriving}
Chen, C., Seff, A., Kornhauser, A., Xiao, J.: Deepdriving: Learning affordance
  for direct perception in autonomous driving. In: Proceedings of the 2015 IEEE
  International Conference on Computer Vision (ICCV). pp. 2722--2730. ICCV '15,
  IEEE Computer Society, Washington, DC, USA (2015).
  \doi{10.1109/ICCV.2015.312}, \url{http://dx.doi.org/10.1109/ICCV.2015.312}

\bibitem{DosovitskiyK16}
Dosovitskiy, A., Koltun, V.: Learning to act by predicting the future. vol.
  abs/1611.01779 (2017), \url{http://arxiv.org/abs/1611.01779}

\bibitem{carla}
Dosovitskiy, A., Ros, G., Codevilla, F., L\'{o}pez, A., Koltun, V.: {CARLA}: An
  open urban driving simulator. In: Conference on Robot Learning (CoRL) (2017)

\bibitem{Frrr2016}
Furrer, F., Burri, M., Achtelik, M., Siegwart, R.: {R}otor{S}{\textemdash}{A}
  modular gazebo {M}{A}{V} simulator framework, Studies in Computational
  Intelligence, vol.~625. Springer, Cham (2016)

\bibitem{gaidon2016virtual}
Gaidon, A., Wang, Q., Cabon, Y., Vig, E.: Virtual worlds as proxy for
  multi-object tracking analysis. In: Proceedings of the IEEE Conference on
  Computer Vision and Pattern Recognition. pp. 4340--4349 (2016)

\bibitem{Atari}
Guo, X., Singh, S., Lee, H., Lewis, R., Wang, X.: Deep learning for real-time
  atari game play using offline monte-carlo tree search planning. In:
  Proceedings of the 27th International Conference on Neural Information
  Processing Systems. pp. 3338--3346. NIPS'14, MIT Press, Cambridge, MA, USA
  (2014), \url{http://dl.acm.org/citation.cfm?id=2969033.2969199}

\bibitem{parkourSimulation}
Ha, S., Liu, C.K.: Iterative training of dynamic skills inspired by human
  coaching techniques. ACM Trans. Graph.  \textbf{34}(1),  1:1--1:11 (Dec
  2014). \doi{10.1145/2682626}, \url{http://doi.acm.org/10.1145/2682626}

\bibitem{unity3Dphysics}
Hamalainen, P., Eriksson, S., Tanskanen, E., Kyrki, V., Lehtinen, J.: Online
  motion synthesis using sequential monte carlo. ACM Trans. Graph.
  \textbf{33}(4),  51:1--51:12 (Jul 2014). \doi{10.1145/2601097.2601218},
  \url{http://doi.acm.org/10.1145/2601097.2601218}

\bibitem{balancingSimulator}
Hamalainen, P., Rajamaki, J., Liu, C.K.: Online control of simulated humanoids
  using particle belief propagation. ACM Trans. Graph.  \textbf{34}(4),
  81:1--81:13 (Jul 2015). \doi{10.1145/2767002},
  \url{http://doi.acm.org/10.1145/2767002}

\bibitem{syntheticCarRecognition}
Hejrati, M., Ramanan, D.: Analysis by synthesis: 3d object recognition by
  object reconstruction. In: Computer Vision and Pattern Recognition (CVPR),
  2014 IEEE Conference on. pp. 2449--2456 (June 2014).
  \doi{10.1109/CVPR.2014.314}

\bibitem{Hussein:2017:ILS:3071073.3054912}
Hussein, A., Gaber, M.M., Elyan, E., Jayne, C.: Imitation learning: A survey of
  learning methods. ACM Comput. Surv.  \textbf{50}(2),  21:1--21:35 (Apr 2017).
  \doi{10.1145/3054912}, \url{http://doi.acm.org/10.1145/3054912}

\bibitem{birdFlightSimulator}
Ju, E., Won, J., Lee, J., Choi, B., Noh, J., Choi, M.G.: Data-driven control of
  flapping flight. ACM Trans. Graph.  \textbf{32}(5),  151:1--151:12 (Oct
  2013). \doi{10.1145/2516971.2516976},
  \url{http://doi.acm.org/10.1145/2516971.2516976}

\bibitem{Kim2015DeepNN}
Kim, D.K., Chen, T.: Deep neural network for real-time autonomous indoor
  navigation. CoRR  \textbf{abs/1511.04668} (2015)

\bibitem{Koutnik:2013}
Koutn\'{\i}k, J., Cuccu, G., Schmidhuber, J., Gomez, F.: Evolving large-scale
  neural networks for vision-based reinforcement learning. In: Proceedings of
  the 15th Annual Conference on Genetic and Evolutionary Computation. pp.
  1061--1068. GECCO '13, ACM, New York, NY, USA (2013).
  \doi{10.1145/2463372.2463509},
  \url{http://doi.acm.org/10.1145/2463372.2463509}

\bibitem{Koutník2014}
Koutn{\'i}k, J., Schmidhuber, J., Gomez, F.: Online Evolution of Deep
  Convolutional Network for Vision-Based Reinforcement Learning, pp. 260--269.
  Springer International Publishing, Cham (2014).
  \doi{10.1007/978-3-319-08864-8\_25}

\bibitem{UE4simulator}
Lerer, A., Gross, S., Fergus, R.: {L}earning {P}hysical {I}ntuition of {B}lock
  {T}owers by {E}xample (2016), arXiv:1603.01312v1

\bibitem{guidedpolicysearch}
Levine, S., Koltun, V.: Guided policy search. In: Dasgupta, S., McAllester, D.
  (eds.) Proceedings of the 30th International Conference on Machine Learning.
  Proceedings of Machine Learning Research, vol.~28, pp.~1--9. PMLR, Atlanta,
  Georgia, USA (17--19 Jun 2013),
  \url{http://proceedings.mlr.press/v28/levine13.html}

\bibitem{deepReinforcementSimulator}
Lillicrap, T.P., Hunt, J.J., Pritzel, A., Heess, N., Erez, T., Tassa, Y.,
  Silver, D., Wierstra, D.: Continuous control with deep reinforcement
  learning. ICLR  \textbf{abs/1509.02971} (2016),
  \url{http://arxiv.org/abs/1509.02971}

\bibitem{loquercio2018dronet}
Loquercio, A., Maqueda, A.I., del Blanco, C.R., Scaramuzza, D.: Dronet:
  Learning to fly by driving. IEEE Robotics and Automation Letters
  \textbf{3}(2),  1088--1095 (2018)

\bibitem{Pedestrian2010}
Mar\'{i}n, J., V\'{a}zquez, D., Ger\'{o}nimo, D., L\'{o}pez, A.M.: Learning
  appearance in virtual scenarios for pedestrian detection. In: 2010 IEEE
  Computer Society Conference on Computer Vision and Pattern Recognition. pp.
  137--144 (June 2010). \doi{10.1109/CVPR.2010.5540218}

\bibitem{mnih2016asynchronous}
Mnih, V., Badia, A.P., Mirza, M., Graves, A., Lillicrap, T., Harley, T.,
  Silver, D., Kavukcuoglu, K.: Asynchronous methods for deep reinforcement
  learning. In: International Conference on Machine Learning. pp. 1928--1937
  (2016)

\bibitem{AtariNature}
Mnih, V., Kavukcuoglu, K., Silver, D., Rusu, A.A., Veness, J., Bellemare, M.G.,
  Graves, A., Riedmiller, M., Fidjeland, A.K., Ostrovski, G., Petersen, S.,
  Beattie, C., Sadik, A., Antonoglou, I., King, H., Kumaran, D., Wierstra, D.,
  Legg, S., Hassabis, D.: Human-level control through deep reinforcement
  learning. Nature  \textbf{518}(7540),  529--533 (Feb 2015),
  \url{http://dx.doi.org/10.1038/nature14236}

\bibitem{syntheticVehicleTraining}
Movshovitz-Attias, Y., Sheikh, Y., Naresh~Boddeti, V., Wei, Z.: 3d
  pose-by-detection of vehicles via discriminatively reduced ensembles of
  correlation filters. In: Proceedings of the British Machine Vision
  Conference. BMVA Press (2014). \doi{http://dx.doi.org/10.5244/C.28.53}

\bibitem{Mueller2016}
Mueller, M., Smith, N., Ghanem, B.: A Benchmark and Simulator for UAV Tracking,
  pp. 445--461. Springer International Publishing, Cham (2016).
  \doi{10.1007/978-3-319-46448-0\_27},
  \url{http://dx.doi.org/10.1007/978-3-319-46448-0\_27}

\bibitem{sim4cv}
M\"{u}ller, M., Casser, V., Lahoud, J., Smith, N., Ghanem, B.: Sim4cv: A
  photo-realistic simulator for computer vision applications. Int. J. Comput.
  Vision  \textbf{126}(9),  902--919 (Sep 2018).
  \doi{10.1007/s11263-018-1073-7},
  \url{https://doi.org/10.1007/s11263-018-1073-7}

\bibitem{NIPS2005}
Muller, U., Ben, J., Cosatto, E., Flepp, B., Cun, Y.L.: Off-road obstacle
  avoidance through end-to-end learning. In: Weiss, Y., Sch\"{o}lkopf, P.B.,
  Platt, J.C. (eds.) Advances in Neural Information Processing Systems 18, pp.
  739--746. MIT Press (2006),
  \url{http://papers.nips.cc/paper/2847-off-road-obstacle-avoidance-through-end-to-end-learning.pdf}

\bibitem{tx1benchmark}
Nvidia: Gpu-based deep learning inference: A performance and power analysis
  (November 2015),
  \url{https://www.nvidia.com/content/tegra/embedded-systems/pdf/jetson\_tx1\_whitepaper.pdf}

\bibitem{syntheticRGBD}
Papon, J., Schoeler, M.: Semantic pose using deep networks trained on synthetic
  {RGB-D}. CoRR  \textbf{abs/1508.00835} (2015),
  \url{http://arxiv.org/abs/1508.00835}

\bibitem{2016-TOG-deepRL}
Peng, X.B., Berseth, G., van~de Panne, M.: Terrain-adaptive locomotion skills
  using deep reinforcement learning. ACM Transactions on Graphics (Proc.
  SIGGRAPH 2016)  \textbf{35}(4) (2016)

\bibitem{2017-TOG-deepLoco}
Peng, X.B., Berseth, G., Yin, K., van~de Panne, M.: Deeploco: Dynamic
  locomotion skills using hierarchical deep reinforcement learning. ACM
  Transactions on Graphics (Proc. SIGGRAPH 2017)  \textbf{36}(4) (2017)

\bibitem{teaching3D}
Pepik, B., Stark, M., Gehler, P., Schiele, B.: Teaching 3d geometry to
  deformable part models. In: Computer Vision and Pattern Recognition (CVPR),
  2012 IEEE Conference on. pp. 3362--3369 (June 2012).
  \doi{10.1109/CVPR.2012.6248075}

\bibitem{pomerleau1989alvinn}
Pomerleau, D.A.: Advances in neural information processing systems 1. chap.
  ALVINN: An Autonomous Land Vehicle in a Neural Network, pp. 305--313. Morgan
  Kaufmann Publishers Inc., San Francisco, CA, USA (1989),
  \url{http://dl.acm.org/citation.cfm?id=89851.89891}

\bibitem{uavHIL2015}
Prabowo, Y.A., Trilaksono, B.R., Triputra, F.R.: Hardware in{-}the{-}loop
  simulation for visual servoing of fixed wing uav. In: Electrical Engineering
  and Informatics (ICEEI), 2015 International Conference on. pp. 247--252 (Aug
  2015). \doi{10.1109/ICEEI.2015.7352505}

\bibitem{GtaV}
Richter, S.R., Vineet, V., Roth, S., Koltun, V.: Playing for data: Ground truth
  from computer games. In: ECCV (2016)

\bibitem{Dagger}
Ross, S., Gordon, G.J., Bagnell, J.A.: No-regret reductions for imitation
  learning and structured prediction. CoRR  \textbf{abs/1011.0686} (2010),
  \url{http://arxiv.org/abs/1011.0686}

\bibitem{Sadeghi2017}
Sadeghi, F., Levine, S.: {CAD2RL}: Real single-image flight without a single
  real image (2017)

\bibitem{AirSim}
Shah, S., Dey, D., Lovett, C., Kapoor, A.: Airsim: High-fidelity visual and
  physical simulation for autonomous vehicles (2017)

\bibitem{Shah:2016}
Shah, U., Khawad, R., Krishna, K.M.: Deepfly: Towards complete autonomous
  navigation of mavs with monocular camera. In: Proceedings of the Tenth Indian
  Conference on Computer Vision, Graphics and Image Processing. pp. 59:1--59:8.
  ICVGIP '16, ACM, New York, NY, USA (2016). \doi{10.1145/3009977.3010047},
  \url{http://doi.acm.org/10.1145/3009977.3010047}

\bibitem{ForestTrail}
{Smolyanskiy}, N., {Kamenev}, A., {Smith}, J., {Birchfield}, S.: {Toward
  Low-Flying Autonomous MAV Trail Navigation using Deep Neural Networks for
  Environmental Awareness}. ArXiv e-prints  (May 2017)

\bibitem{bikeStunts}
Tan, J., Gu, Y., Liu, C.K., Turk, G.: Learning bicycle stunts. ACM Trans.
  Graph.  \textbf{33}(4),  50:1--50:12 (Jul 2014).
  \doi{10.1145/2601097.2601121},
  \url{http://doi.acm.org/10.1145/2601097.2601121}

\bibitem{hilUAV}
Trilaksono, B.R., Triadhitama, R., Adiprawita, W., Wibowo, A., Sreenatha, A.:
  Hardware{-}in{-}the{-}loop simulation for visual target tracking of octorotor
  uav. Aircraft Engineering and Aerospace Technology  \textbf{83}(6),  407--419
  (2011). \doi{10.1108/00022661111173289},
  \url{http://dx.doi.org/10.1108/00022661111173289}

\bibitem{torcs}
Wymann, B., Dimitrakakis, C., Sumner, A., Espi\'e, E., Guionneau, C., Coulom,
  R.: {TORCS}, the open racing car simulator. \texttt{http://www.torcs.org}
  (2014)

\end{thebibliography}
	
\end{document}